\documentclass[12pt, reqno]{amsart}
\usepackage{amsmath, amsthm, amscd, amsfonts, amssymb, graphicx, color}
\usepackage[bookmarksnumbered, colorlinks, plainpages]{hyperref}

\theoremstyle{definition}

\theoremstyle{remark}

\numberwithin{equation}{section}

\begin{document}
\setcounter{page}{1}

\noindent {\small \textit{To appear on arXiv}} \hfill {\small \texttt{arXiv:2507.14560 [cs.LG]}}\\
{\small Version 1, July 2025}

\centerline{}

\centerline{}

%------------------------------------------------------------------------------

%Title of the paper
\title[The Roots of Self-Attention]{The Origin of Self-Attention: Pairwise Affinity Matrices in Feature Selection and the Emergence of Self-Attention}

%Author names and affiliations
\author[Affinity-Based Computation]{*Giorgio Roffo}

%In case of 3 or more authors use below format
%\author[F. Author, S. Author, T. Author]{First Author$^1$, Second Author$^2$$^{*}$ \MakeLowercase {and} Third Author$^3$}

\email{\textcolor[rgb]{0.00,0.00,0.84}{giorgio.roffo@gmail.com}}

% Optional second author
%\address{$^{2}$ Department of Mathematics, University of California, San Diego, USA.}
%\email{\textcolor[rgb]{0.00,0.00,0.84}{second.author@ucsd.edu}}

%\dedicatory{This paper is dedicated to Professor ABCD}

\date{Received: July 2025.
\newline \indent $^{*}$ Corresponding author
\newline \indent © The Author(s) 2025. This work is licensed under a Creative Commons Attribution-
\newline \indent NonCommercial-NoDerivatives 4.0 International License. To view a copy of the licence, visit
\newline \indent \url{https://creativecommons.org/licenses/by-nc-nd/4.0/}}

%Abstract, keywords, math subject classification
% Abstract, keywords, math subject classification
\begin{abstract}
The self-attention mechanism, now central to deep learning architectures such as Transformers, represents a modern instantiation of a broader computational principle: learning and leveraging pairwise affinity matrices to modulate information flow. This paper traces the conceptual lineage of self-attention across domains—vision, language, and graph learning—through the shared structure of an affinity matrix $A$. In particular, we spotlight \textit{Infinite Feature Selection} (Inf-FS) as a foundational framework that generalizes the notion of affinity-based weighting. Unlike the fixed dot-product formulation in Transformers, Inf-FS defines $A$ flexibly—either handcrafted or learned—and computes feature importance through multi-hop propagation over the affinity graph. Methodologically, this positions Inf-FS as a superset: self-attention arises as a specific case where $A$ is parameterized via learned token similarities and applied in a single-hop fashion. We argue that the core structure—reasoning over pairwise relationships—is conserved, and the main distinction lies in how $A$ is constructed and utilized. By reframing self-attention within the broader paradigm of affinity-based computation, we unify disparate threads in machine learning and underscore a shared mathematical foundation that transcends task or architecture.

\end{abstract}
\maketitle

%------------------------------------------------------------------------------
\section{Introduction and Preliminaries}

A pairwise affinity matrix \( A \in \mathbb{R}^{N \times N} \) represents relationships or similarities between elements \( i, j \in \{1, \dots, N\} \). Such matrices are fundamental in pattern recognition, spectral methods, and graph-based learning, where the structure of interactions among elements is essential for modeling global context \cite{vonLuxburg2007, Newman2003, Zhou2004}.

Affinity matrices have long been used in machine learning and pattern recognition to represent graphs or networks of interactions. Early examples include graph adjacency matrices in spectral clustering (where \( A_{ij} \) might be a Gaussian similarity between data points) \cite{Ng2002, vonLuxburg2007} and social network graphs for centrality analysis \cite{Newman2003, Bonacich1987}.

In this work, we trace the origin of self-attention to a broader tradition of affinity-based reasoning. In particular, we highlight Infinite Feature Selection (Inf-FS), introduced in 2015, as a pivotal development that established the use of a fully connected affinity matrix \( A \) to quantify and propagate feature importance \cite{Roffo2015}. Inf-FS provides a general-purpose framework where \( A \) can be defined via statistical, semantic, or learned measures, and is used to:
\begin{enumerate}
    \item \textbf{Score} features based on their connectedness;
    \item \textbf{Weight} features according to their importance;
    \item \textbf{Rank} features globally across a dataset;
    \item \textbf{Select} informative subsets of features.
\end{enumerate}

The key idea is that by comparing all pairs of elements, one can capture global structure or importance. The formulation is mathematically grounded in a power series expansion over \( A \), which aggregates contributions from all possible paths of interaction between features—direct and indirect. This approach anticipated the key structural idea of self-attention: that relationships between input elements can be encoded in a pairwise affinity matrix and used to control the flow of information \cite{Roffo2015}.

While Inf-FS was developed in the context of feature selection, the same structure, an affinity matrix \( A \) used for weighted aggregation, appeared later in self-attention mechanisms, particularly in models where input elements (e.g., tokens, patches, nodes) are reweighted based on their contextual relevance. In this view, self-attention emerges as a special case where \( A \) is constructed dynamically from learned input representations and applied in a single-hop update \cite{Vaswani2017}.

Self-attention (also called intra-attention) refers to an attention mechanism that relates a sequence or set with itself, as opposed to the earlier “attention” in sequence-to-sequence models which related one sequence with another (e.g., decoder attending to encoder states in machine translation \cite{Bahdanau2015}). Importantly, self-attention is implemented via an affinity matrix \( A \) of size \( N \times N \) (for \( N \) input elements) that effectively forms a fully-connected graph over the elements. Each \( A_{ij} \) indicates how much element \( i \) should pay attention to element \( j \). This structural idea—computing pairwise comparisons to derive contextual weights—originated in the domain of feature selection, particularly within the re-weighting step that precedes the selection of informative features \cite{Roffo2015, Roffo2021, Gui2019}. In this setting, re-weighting serves to estimate the relative importance of each feature in relation to all others, based on affinity-based interactions. Although this process was not originally framed as “attention,” it anticipates the same mechanism of importance-driven weighting found in self-attention.

Below, we trace the lineage of self-attention through the lens of affinity matrices, covering developments in feature selection, natural language processing, computer vision, and graph neural networks. We highlight foundational works that introduced or relied on pairwise affinity structures, and show how they conceptually connect to the modern Transformer-style attention (where $A = QK^\top$ is the core).

\textit{This paper does not claim that Inf-FS introduced self-attention in its modern form}. Rather, it emphasizes that Inf-FS introduced a general mathematical structure that was later adopted and operationalized in neural self-attention. Several later works, including AFS \cite{Gui2019AFS} and Sequential Attention \cite{Zhao2023Sequential}, explicitly use attention weights for feature selection, further bridging these ideas.

We now present a historical timeline highlighting key developments in affinity-based reasoning across domains, culminating in the adoption of self-attention as a central mechanism in deep learning.

\section*{Timeline of Key Developments in Affinity-Based Attention}

The following timeline outlines major developments in the use of pairwise affinity matrices across disciplines, with an emphasis on their role in feature selection, attention mechanisms, and related learning frameworks.

\begin{itemize}

    \item \textbf{1950s–1960s (Psychological Foundations)}: Early theories of human attention emerged in cognitive psychology and neuroscience, including the cocktail party effect \cite{Cherry1953}, Broadbent’s filter model \cite{Broadbent1958}, and Sperling’s partial report paradigm \cite{Sperling1960}. These works motivated later computational models of attention.

    \item \textbf{1980s–1990s (Precursors to Computational Attention)}: Neural architectures incorporating multiplicative interactions, such as sigma-pi units \cite{Rumelhart1986}, higher-order neural networks \cite{Giles1987}, and fast weight controllers \cite{Schmidhuber1992}, laid the groundwork for key-value mechanisms in attention.

    \item \textbf{1998–2005 (Affinity in Image Processing and Graph Ranking)}: 
    Bilateral filtering \cite{Tomasi1998} and non-local means \cite{Buades2005} used Gaussian affinity functions to smooth images based on patch similarity. Concurrently, the PageRank algorithm \cite{Page1998} ranked nodes in web graphs via random walks, effectively diffusing relevance through a pairwise affinity matrix. These methods demonstrated the utility of global interactions encoded via pairwise similarities, despite being non-learnable.

    \item \textbf{2014 (Learned Attention in Feature Selection)}: 
    Wang et al. introduced the “Attentional Neural Network” \cite{Wang2014}, incorporating top-down cognitive modulation for feature weighting in noisy visual recognition tasks. This was an early example of attention-inspired reweighting applied to feature selection.

    \item \textbf{2015 (Infinite Feature Selection)}: 
    Roffo et al. proposed Infinite Feature Selection (Inf-FS) \cite{Roffo2015}, introducing a fully connected affinity matrix \( A \) between features to model pairwise relationships. A ranking score is computed via a power series expansion:
    \[
        S = \sum_{k=1}^{\infty} \alpha^k A^k = (I - \alpha A)^{-1} - I, \quad \text{for } 0 < \alpha < \frac{1}{\rho(A)}
    \]
    Inf-FS formalized feature scoring, weighting, ranking, and selection through global affinity propagation. It did not frame the approach as “attention,” but it anticipated the structural form later used in self-attention mechanisms.

    Bahdanau et al. introduced attention for sequence-to-sequence models in neural machine translation \cite{Bahdanau2015}. Alignment scores were computed between encoder and decoder states, normalized via softmax to reweight source representations.

    \item \textbf{2016–2017 (Self-Attention in Recurrent Models)}: 
    Attention was extended within sequences via self-attention modules. Cheng et al. \cite{Cheng2016}, Parikh et al. \cite{Parikh2016}, and Lin et al. \cite{Lin2017} applied intra-sequence attention for reading comprehension and sentence encoding, replacing or augmenting recurrent connections.

    \item \textbf{2017 (Transformer Architecture)}: 
    Vaswani et al. introduced the Transformer \cite{Vaswani2017}, which eliminated recurrence and adopted scaled dot-product self-attention as the core mechanism. Each layer computes:
    \[
        \mathrm{Attention}(Q, K, V) = \mathrm{softmax}\left(\frac{QK^\top}{\sqrt{d_k}}\right)V
    \]
    This generalized the idea of contextual weighting via affinity matrices to end-to-end learning over sequences.

    \item \textbf{2017–2019 (Pairwise Reasoning in Sets and Graphs)}: 
    Relation Networks \cite{Santoro2017} modeled interactions over object pairs for relational reasoning. Set Transformers \cite{Lee2019} used permutation-invariant self-attention for unordered sets. Graph Attention Networks (GAT) \cite{Velickovic2018} introduced learnable attention over graph edges.

    \item \textbf{2018 (Non-local Attention in Vision)}: 
    Wang et al. proposed Non-Local Neural Networks \cite{Wang2018NonLocal}, applying attention to image and video features by computing long-range dependencies via learned similarity functions, paralleling the Transformer formulation.

    \item \textbf{2019–2020 (Efficient Attention)}: 
    Several methods addressed the scalability of attention for long sequences. These include Reformer \cite{Kitaev2020}, Linformer \cite{Wang2020Linformer}, Performer \cite{Choromanski2021}, and Efficient Attention \cite{Shen2021Efficient}. These variants introduced structural or kernel-based approximations to reduce computational cost while preserving pairwise weighting.

    \item \textbf{2021 (Inf-FS and Structural Convergence)}: 
    Ramsauer et al. \cite{Ramsauer2021} formalized connections between Hopfield networks and Transformers, interpreting attention as associative retrieval over affinity matrices.

    \item \textbf{2022–2023 (Unified Affinity-Based Modeling)}: 
    Self-attention mechanisms have since been adopted in vision (e.g., Vision Transformers), biology (e.g., AlphaFold), and vision-language models (e.g., CLIP). At the same time, several works applied attention weights to feature selection tasks explicitly, such as AFS \cite{Gui2019AFS} and Sequential Attention \cite{Zhao2023Sequential}, reinforcing the continuity with the Inf-FS framework.
    
\end{itemize}

Although Inf-FS and self-attention were developed in different contexts—feature selection and sequence modeling, respectively—both frameworks employ a common structural element: a pairwise affinity matrix \( A \in \mathbb{R}^{N \times N} \) that encodes relationships among input elements.

In Inf-FS, the matrix \( A \) is used to model affinities between features and may be defined analytically (e.g., using correlation or distance metrics) or learned from data. The method ranks features by aggregating relevance scores over multi-hop paths on the induced affinity graph, where paths of arbitrary length are considered via a matrix power series. In contrast, self-attention computes \( A \) dynamically from learned token embeddings, typically using a scaled dot-product formulation, and applies the resulting weights in a single-hop aggregation. Deeper interactions in self-attention arise through layer stacking rather than explicit multi-hop propagation.

Methodologically, Inf-FS provides a general framework for scoring and selecting elements based on affinity-based propagation. Self-attention aligns with this framework when the affinity matrix is constructed from data-dependent similarities and used to weight contextual contributions. The distinction lies in the construction and application of \( A \): Inf-FS emphasizes global feature relevance across the input space, while self-attention focuses on local context within a specific forward pass.

This comparison highlights the structural similarity between the two approaches and illustrates how a common mathematical mechanism—pairwise affinity modeling—can be instantiated under different inductive biases and computational constraints.

\section{Graph-Based Feature Selection and Early Affinity Approaches}

One of the earliest domains to explicitly use an affinity matrix for weighting importance was feature selection in machine learning. Feature selection aims to identify which input features (variables) are most relevant for a task, often by scoring or ranking features. Traditional filter methods scored features individually (e.g., by correlation or mutual information with the target), ignoring feature–feature interactions. In the mid-2010s, researchers began formulating feature selection as a graph problem to capture feature interactions.

Infinite Feature Selection (Inf-FS) by Roffo et al. \cite{InfFS2015} is a seminal work that introduced a fully-connected feature graph approach. In Inf-FS, each feature is a node in a graph, and an edge between feature $i$ and $j$ is weighted by an affinity score reflecting how related or redundant those two features are. For example, one can define $A_{ij}$ based on statistical measures (the paper uses a mix of feature correlation and variance). This yields an affinity matrix $A$ where features are compared pairwise. Rather than selecting features in isolation, Inf-FS evaluates each feature in the context of all other features, which was a departure from earlier methods.

Crucially, Inf-FS introduced the concept of considering paths of any length in the feature graph as feature subsets. It leverages the convergence of a power series of matrices: essentially summing $A + A^2 + A^3 + \cdots$ to infinity (with appropriate normalization for convergence). Intuitively, $A^2$ captures two-hop relationships (feature $i$ connected to $k$ via some intermediate feature $j$), $A^3$ captures three-hop paths, and so on. By summing these, Inf-FS obtains a score for each feature that accounts for all possible interaction paths among features. A feature that is strongly connected to others (either directly or through chains) will have a higher score, meaning it’s either highly relevant or acts as a hub that connects clusters of features. This procedure yields a feature ranking, hence performing selection by choosing top-ranked features.

Mathematically, if $S = A + A^2 + A^3 + \cdots$, one can show $S = (I - A)^{-1} - I$ (assuming $\|A\|<1$ for convergence). The $i$-th row/column sum of $S$ (or another aggregation of $S$) can serve as the importance score of feature $i$. In practice, Roffo et al. introduced a parameter $\alpha$ to weight the influence of longer paths (a kind of decay), but the core idea is summing infinitely many walk contributions. Because it conceptually allows paths of unbounded length, they called it “infinite” feature selection.

Why is this relevant to self-attention? It turns out that if we limit Inf-FS to path length 1 only, it uses just the matrix $A$ itself (direct feature affinities) to score features. This degenerate case means each feature’s score is simply the sum of its edge weights to all other features – essentially a weighted degree centrality. That is structurally analogous to what a single self-attention layer does: it uses the affinity matrix (after normalization) to compute a weighted combination of features. In fact, one can see self-attention (single layer) as performing a one-hop aggregation on a fully-connected graph of tokens. Inf-FS with full infinite paths goes beyond one-hop (more akin to stacking multiple attention layers or doing a power series expansion in one go), but the one-hop formulation is the same graph operation underlying self-attention. The Inf-FS authors explicitly note that their method considers “a subset of features as a path connecting them” and uses the affinity graph to evaluate feature relevance – a description that closely parallels how attention connects tokens in a sequence via a fully-connected graph of affinities.

Inf-FS was a key milestone, but it built on earlier ideas of graph centrality for features. For instance, Roffo and colleagues also explored Eigenvector Centrality Feature Selection (EC-FS), in which features are ranked by their principal eigenvector centrality in a feature affinity graph (this effectively uses $A$ to score features via the eigenvector equation $Av=\lambda v$). Such approaches treat highly interconnected features as important. Another related idea is using PageRank on a feature graph to score features (drawing an analogy to webpages “voting” for each other); in feature selection, this was sometimes used to diffuse importance across a feature network. These methods all share the notion of an affinity matrix among features. Inf-FS distinguished itself by analytically summing all paths instead of truncating at a fixed length or relying on an iterative eigenvector solution, thereby theoretically considering all higher-order interactions among features.

Extensions of Inf-FS reinforced its connection to learnable affinity matrices. In ICCV 2017, Roffo et al. introduced Infinite Latent Feature Selection \cite{InfLatentFS2017}. This method retained the idea of a fully-connected feature graph and infinite path ranking, but instead of defining edge weights by a simple fixed function (correlation, etc.), it learned the affinities using a latent variable model. Specifically, they used a probabilistic latent semantic analysis (PLSA)-inspired approach to learn edge weights that reflect the probability that two features jointly indicate an underlying “relevancy” factor. In effect, the graph became trainable: the better it explained the data in terms of a latent notion of feature relevancy, the higher the edge weights between co-relevant features. The ranking of features was then done on this learned graph. This was an important step toward making the affinity matrix $A$ data-driven, foreshadowing how self-attention learns affinities on the fly from input data. A later journal version (TPAMI 2020) of Inf-FS \cite{InfFS2021} further solidified the approach and even listed “feature reweighting” and “attention” as keywords, explicitly acknowledging that the Inf-FS mechanism can be seen as an attention-like operation over features.

In summary, by the time of 2015–2017, the idea of computing a pairwise relevance matrix $A$ among input variables and using it to weight or select features was well-established in feature selection research. This represents one lineage of self-attention: treating input features as nodes in a graph and using their mutual affinities to decide importance. The difference was that feature selection methods produced a static ranking or mask (often not input-specific, or computed on the entire dataset), whereas self-attention would soon produce dynamic weightings per input instance. Nonetheless, the structural similarity is clear: Inf-FS built a fully-connected graph of features exactly as a Transformer builds a fully-connected graph of tokens.

\section*{Emergence of Self-Attention in Sequence Models (NLP)}

In natural language processing (NLP), the notion of “attention” first rose to prominence in the context of Neural Machine Translation. The landmark work by Bahdanau, Cho, and Bengio (2015) introduced an attention mechanism in an encoder–decoder RNN that allowed the decoder to focus on different parts of the source sentence when producing each word of the translation \cite{Bahdanau2015}. This was a cross-attention: at each decoder timestep, a weight was computed for each encoder hidden state (each source word) based on a similarity between the decoder’s state and the encoder state. Those weights (forming a vector that sums to 1) served to create a weighted sum of encoder representations – essentially a dynamic context vector for the decoder. Although not self-attention (since it connects two sequences), this introduced the key idea of using learned compatibilities (affinities) to weight information. It demonstrated the power of letting a model learn where to look by comparing representations (query vs. keys in modern terms) and weighting accordingly.

The step toward self-attention (within one sequence) came soon after. Researchers realized you could apply similar mechanisms to let different positions of the same sequence attend to each other, enhancing how much context each position’s representation can capture. For example, Cheng, Dong, and Lapata (2016) implemented a form of self-attention in a reading comprehension model \cite{Cheng2016}. They modified a bi-directional LSTM such that at each time step, the model could attend to all previous words (a sort of memory) instead of just relying on the recurrent state. Similarly, Paulus et al. (2017) and others in text summarization used “intra-attention” where the decoder attends over its own previously generated outputs to avoid repetition. Another notable work is Lin et al. (ICLR 2017), who proposed a self-attentive sentence embedding: they computed multiple attention weight vectors over the tokens of a sentence (using a learned parameter matrix) to extract a richer sentence representation \cite{Lin2017}. All these can be seen as precursors that within an RNN framework (or alongside it) introduced learnable affinity-based weighting among tokens.

The culmination of these ideas was the Transformer model by Vaswani et al. (NeurIPS 2017), famously introduced in the paper “Attention Is All You Need.” The Transformer did away with recurrent networks entirely and relied solely on self-attention mechanisms to encode sequences \cite{Vaswani2017}. In a Transformer layer (encoder or decoder), each position $i$ attends to all positions $j$ in the same sequence (or to all positions in the source sequence, for decoder cross-attention) through a learned affinity matrix.

\textbf{Scaled Dot-Product Self-Attention:} Formally, the Transformer computes for each position $i$ a weighted sum of the input representations (value vectors) at all positions, using weights derived from pairwise dot-products. Each input token $i$ is associated with a query vector $q_i$ and each token $j$ with a key vector $k_j$ (these come from learned linear projections of the token’s embedding or the previous layer’s output). The unnormalized attention score from $i$ to $j$ is the dot product $q_i \cdot k_j$ (how similar token $i$’s query is to token $j$’s key). In matrix form, if $Q$ is the matrix of all query vectors and $K$ the matrix of all key vectors, the affinity matrix is simply:

\[
A = QK^T,
\]

an $N\times N$ matrix where $A_{ij} = q_i \cdot k_j$. The Transformer then applies a scaling factor (dividing by $\sqrt{d_k}$, the dimensionality of $q_i$) and a row-wise softmax. Let $\tilde{A}_{ij} = \frac{q_i \cdot k_j}{\sqrt{d_k}}$. Then the attention weight matrix is:

\[
W = \text{softmax}(\tilde{A}),
\]

where $W_{ij} = \frac{\exp(\tilde{A}_{ij})}{\sum_{m=1}^N \exp(\tilde{A}_{im})}$. Each row of $W$ sums to 1, and $W_{ij}$ represents how much attention token $i$ pays to token $j$. Finally, these weights are used to take a weighted combination of value vectors $v_j$ (another projection of token $j$). The output for token $i$ is $z_i = \sum_{j} W_{ij} v_j$. In matrix form:

\[
\text{Attention}(Q,K,V) = \text{softmax}\left(\frac{QK^T}{\sqrt{d_k}}\right)V.
\]

This equation (often referred to as “scaled dot-product attention”) makes the affinity matrix $A=QK^T$ the centerpiece: it encodes all pairwise token similarities, and after normalization, it is used to mix token information. Notably, this is done in parallel for all tokens by efficient matrix operations \cite{Vaswani2017}. The result $Z = AV$ (after softmax) is effectively a new set of token representations where each token $i$ has integrated information from every other token, weighted by their learned affinity \cite{Vaswani2017}. The Transformer also introduced multi-head attention, which means it computes multiple different affinity matrices (with different learned projection subspaces) and multiple weighted sums, then concatenates them \cite{Vaswani2017,Lin2017}. This improves the model’s capacity to capture different types of relationships (for example, one head might attend based on semantic similarity, another on positional patterns, etc.). But the principle remains: compute pairwise affinities and use them to transfer information among all pairs.

\textbf{Advantages of self-attention:} As highlighted by Vaswani et al., a self-attention layer can draw global dependencies in one step – any token can potentially attend to any other token with just one matrix multiplication, regardless of their distance \cite{Vaswani2017}. In contrast, a recurrent network would require many time steps to carry information over long distances. This global receptive field of the affinity matrix is a direct analog of the fully-connected feature graph in Inf-FS, except now it’s fully input-dependent and learned for the task at hand. Self-attention also allows parallel computation over all token pairs (making it computationally attractive for modern hardware) \cite{Vaswani2017}, at the cost of $O(N^2)$ complexity in sequence length. The trade-off of self-attention is that it averages information (an attention head produces a weighted average of token representations, which Vaswani et al. noted can dilute some information \cite{Vaswani2017}). However, using multiple heads and multiple layers mitigates this. Essentially, stacking self-attention layers is analogous to considering multi-hop interactions: one attention layer = one “hop” (direct affinity), two layers can capture two-step relationships (a token attends to another token after that token has attended to a third, indirectly modeling a 2-hop path), and so on. This is reminiscent of how Inf-FS considered $A^2, A^3$, etc., although Transformers do it via stacking rather than an explicit power series sum.

By late 2017, self-attention as defined by the Transformer had become the de facto standard for sequence modeling in NLP. It was quickly adopted in models like BERT (Devlin et al. 2018) and GPT (Radford et al. 2018+) for language understanding and generation. These models demonstrated that learning an affinity matrix over input tokens and using it to propagate information can yield extremely rich representations and state-of-the-art performance across NLP tasks. The concept of “different positions of a single sequence attending to each other” was now mainstream \cite{Vaswani2017}. It’s worth noting that Vaswani et al. themselves cited prior works that used self-attention in RNNs, acknowledging that the idea had been “used successfully in a variety of tasks” before, such as reading comprehension, summarization, and entailment \cite{Vaswani2017}. What the Transformer did was elevate self-attention to the primary computational block and show it can replace recurrence entirely \cite{Vaswani2017}.

From the perspective of affinity matrices: the Transformer firmly established that learning and using a pairwise affinity matrix $A$ is a powerful general approach for representation learning. It wasn’t an isolated innovation, but rather the crystallization of a concept that had been percolating (the notion of comparing elements to each other) into a simple, widely-applicable form. We will see next how this same concept was independently (or subsequently) explored in other fields like computer vision and graph learning, often drawing direct analogies to the Transformer’s mechanism.

\section*{Self-Attention in Computer Vision: Non-Local Neural Networks and Beyond}

Computer vision (CV) problems, such as image classification or segmentation, traditionally rely heavily on local processing (convolutions capture local pixel neighborhoods). Yet, vision researchers have long recognized the value of global relationships – pixels or regions far apart can still be related (think of an image with repeating patterns, or an object whose parts are distant in pixel space but conceptually connected). A classic CV technique that presaged self-attention is the Non-Local Means filter \cite{Buades2005}, a denoising algorithm that for each pixel takes a weighted average of many other pixels in the image, where the weights are based on patch similarity. Non-local means uses an affinity function (typically a Gaussian of patch distance) to determine how much one pixel’s intensity should contribute to denoising another pixel. In formula, it looks like $y_i = \frac{1}{C(i)}\sum_j f(x_i, x_j) x_j$ where $f(x_i,x_j) = \exp(-|P_i - P_j|^2 / h^2)$ for patches $P_i, P_j$ around pixels $i,j$ and $C(i)$ is a normalizing factor. This is strikingly similar to an attention update: an output value is a weighted sum of other values $x_j$ with weights given by an affinity $f(x_i,x_j)$. The key difference is that in non-local means the weights are a fixed function of pixel intensities (not learned), and the goal is image smoothing, not learned representations. Nonetheless, it introduced vision to the non-local averaging concept.

Fast-forward to 2018: Xiaolong Wang et al. published Non-Local Neural Networks \cite{Wang2018NonLocal}, explicitly inspired by the self-attention idea in NLP and by classical non-local filtering in vision. They proposed a non-local block as a generic module that can be inserted into convolutional neural networks to allow global dependencies to be captured. In a non-local block, the feature map (say of shape $N$ locations by $C$ channels) is treated similarly to a sequence: they compute an affinity between any two locations $i$ and $j$ in the feature map and use it to adjust the features. In the simplest version, they define $f(x_i, x_j) = \theta(x_i)^T \phi(x_j)$, where $x_i$ is the feature vector at position $i$ (like a pixel or a region), and $\theta, \phi$ are learnable linear projections (analogous to queries and keys) \cite{Wang2018NonLocal}. This is exactly the dot-product similarity from self-attention (sometimes they also experimented with Gaussian $f(x_i,x_j) = e^{x_i^T x_j}$ or a concatenation-based function) \cite{Wang2018NonLocal}. They then apply a softmax normalization over $j$ for each $i$ (for the Gaussian version; for the pure dot-product version they normalized by $1/N$) \cite{Wang2018NonLocal}. Finally, they compute an output at $i$ as $y_i = \sum_j f(x_i,x_j),g(x_j)$, where $g(x_j)$ is typically a linear embedding (like the “value” in attention, noted as $W_g x_j$) \cite{Wang2018NonLocal}. They add $y$ back to the input (residual connection) and feed it into the next layer. This non-local operation can be applied to image features (where $i,j$ index spatial positions) or video (indexing spatial-temporal positions) or even more abstract graph-like data.

Wang et al. explicitly draw the parallel to the self-attention in machine translation: they point out that the embedded Gaussian version of their non-local block is essentially identical to the self-attention formula \cite{Wang2018NonLocal}. In their words, “the self-attention module \cite{Vaswani2017} recently presented for machine translation is a special case of non-local operations in the embedded Gaussian version”, where “[for a given $i$], $1/C(x) f(x_i,x_j)$ becomes the softmax computation along $j$” \cite{Wang2018NonLocal}. (Here \cite{Vaswani2017} is referencing Vaswani et al. 2017, and $C(x)$ is the normalization factor.) They even rewrite the self-attention equation in their notation: “$y = \text{softmax}(x^T W_\theta^T W_\phi x),g(x)$, which is the self-attention form in \cite{Vaswani2017}” \cite{Wang2018NonLocal}. This acknowledgment is important: it cements that the CV community saw self-attention not as an NLP-specific trick, but as a generic non-local weighting operation applicable to any domain where you have a set of elements (be it image pixels or video frames) and you want to capture long-range interactions \cite{Wang2018NonLocal}.

By introducing the non-local block, Wang et al. achieved notable improvements in video classification and also boosted performance in tasks like object detection and segmentation when added to backbone CNNs \cite{Wang2018NonLocal}. The block gave CNNs a way to adaptively aggregate information from distant parts of an image or sequence, which standard convolution or recurrent layers struggled with. This was effectively a learned global affinity matrix over image regions. Shortly after, many other vision works incorporated similar ideas: e.g. Criss-Cross Attention \cite{Huang2019} for 2D segmentation computed affinities in row and column stripes to approximate full-image attention with less cost; Dual Attention Networks (DANet) \cite{Fu2019} applied two parallel self-attention modules — one spatial (pixel-to-pixel affinities) and one channel-wise (affinities between feature channels) — to better capture contextual and feature relationships for segmentation. The Vision Transformer (ViT) \cite{Dosovitskiy2020} took the concept to the extreme by directly applying a Transformer architecture to image patches, treating them like tokens, thus using self-attention as the sole mechanism for image recognition (with great success as well). All these are variations on the theme of learned affinity matrices guiding feature combination.

It’s worth noting that in vision, there were also earlier graph-based methods that bear resemblance. For instance, fully-connected CRFs (dense Conditional Random Fields) were used in segmentation to refine outputs by considering pairwise pixel affinities (often using Gaussian kernels) – this is a fixed affinity matrix, not learned, but conceptually similar in modeling long-range connections. Also, the concept of a bilateral filter \cite{Tomasi1998} is similar to non-local means (using intensity/color affinity plus spatial proximity to do weighted averaging). These were not learned or called “attention,” but they show that computer vision had independently arrived at the notion of weighting each element by pairwise functions of all elements.

Summary in CV: The “self-attention” structure of an affinity matrix controlling information flow has become ubiquitous in modern CV as well, though often under names like non-local modules or attention modules. The non-local neural network paper unified the view by saying: we have a generic building block that computes responses as a weighted sum of features at all positions, where weights are a function of pairwise feature relations \cite{Wang2018NonLocal}. That is one-to-one what self-attention does. This greatly expanded the reach of attention mechanisms: from sequences of text to 2D or 3D feature maps, and even to more abstract graphs or sets.

\section*{Graph Neural Networks and Attention Mechanisms}

Graph Neural Networks (GNNs) deal with data that are naturally represented as graphs: nodes connected by edges (with arbitrary topology). Before attention came into play, GNNs like the Graph Convolutional Network (GCN by Kipf \& Welling, 2017) used the adjacency matrix of the graph to propagate information, typically with equal or degree-normalized weights for each neighbor. That is, a node’s new representation might be the average (or a weighted sum with learned scalar weights) of its neighbors’ representations. However, a fixed adjacency can be suboptimal – not all neighbors are equally important, and perhaps not all edges should be treated the same for a given task.

Enter Graph Attention Networks (GAT) by Veličković et al. \cite{Velickovic2018}. GATs brought the self-attention paradigm to graphs, enabling nodes to attend to their neighbors with learned weights. In a GAT layer, for each edge $i \to j$ (where $j$ is in the neighborhood of $i$), an attention coefficient $e_{ij}$ is computed as:
\[
e_{ij} = \text{LeakyReLU} \left( \vec{a}^\top [W \vec{h}_i \, \| \, W \vec{h}_j] \right),
\]
where $\vec{h}_i$ and $\vec{h}_j$ are the input features of node $i$ and $j$, $W$ is a weight matrix (applied to every node’s features), $[\cdot \, \| \, \cdot]$ denotes concatenation, and $\vec{a}$ is a learnable weight vector defining the attention mechanism. In plain terms, they use a small one-layer feedforward network (with weight vector $\vec{a}$ and nonlinearity) to compute a score for the pair of nodes $(i,j)$ based on their feature representations. This $e_{ij}$ is analogous to an unnormalized attention score. They then normalize these across all neighbors of $i$ using a softmax:
\[
\alpha_{ij} = \frac{\exp(e_{ij})}{\sum_{k \in \mathcal{N}(i)} \exp(e_{ik})},
\]
where $\mathcal{N}(i)$ is the neighborhood of node $i$ (typically including $i$ itself if a self-loop is considered). These $\alpha_{ij}$ are the attention weights on edges, effectively forming a (sparse) affinity matrix for the graph – sparse because $\alpha_{ij}$ is computed only for $j$ that are connected to $i$ in the original graph structure. This is masked attention, to respect the graph connectivity. Finally, the node features are updated as:
\[
\vec{h}_i' = \sigma \left( \sum_{j \in \mathcal{N}(i)} \alpha_{ij} W' \vec{h}_j \right),
\]
where $W'$ is another weight matrix (often the same as $W$ or a separate “value” weight) and $\sigma$ is an activation function. This is exactly the same pattern: each node $i$ takes a weighted combination of its neighbors’ feature vectors, with weights $\alpha_{ij}$ that were computed by an attention mechanism. GAT also employed multi-head attention (computing multiple $\alpha^{(head)}_{ij}$ and corresponding outputs and then concatenating or averaging), akin to the Transformer, to stabilize training and enrich capacity.

The GAT paper makes a point that their attentional mechanism is “agnostic to the graph structure” beyond the mask – meaning it doesn’t need to know global graph properties or do expensive spectral operations; it just learns to focus on the most relevant neighbors for each node \cite{Velickovic2018}. This was a big improvement in flexibility and performance on node classification tasks. Importantly, GAT highlighted that attention weights on graph edges improve interpretability: one can inspect $\alpha_{ij}$ to see which neighbors a node is paying most attention to, just like one can analyze which words attend to which in a sentence. This is a general benefit of attention mechanisms – they provide a set of learned affinity weights that can often be interpreted in the context of the data.

From the affinity matrix perspective, GAT shows that even when you have an existing graph structure, it can be beneficial to learn a finer weighting of that graph’s adjacency matrix. The traditional GCN essentially uses a fixed affinity:
\[
\tilde{A}_{ij} = \frac{1}{\sqrt{\deg(i)\deg(j)}}
\]
if $(i,j)$ is an edge (and 0 otherwise). GAT replaces those fixed entries with learned ones $\alpha_{ij}$ (and 0 for non-edges) – a learned, feature-dependent adjacency matrix. In extreme cases, if the graph is fully connected (every node sees every other as neighbor), GAT would learn an affinity matrix much like a dense self-attention (this is rarely done due to computational cost on large graphs, but conceptually possible). In practice, many graph papers following GAT have used attention to learn or refine adjacency. Some even learn graph structure from scratch by starting with no edges or a fully-connected graph but encouraging sparsity, so that the model itself determines which pairwise connections to keep. These can be seen in works on latent graph learning, e.g., “LatentGNN” \cite{LatentGNN2019}, which approximates a full affinity by factorization, or in some attention-based combinatorial optimization solvers.

In parallel, Santoro et al. (NeurIPS 2017) introduced Relation Networks \cite{Santoro2017}, a simple neural module for reasoning that takes a set of “objects” and considers all pairwise interactions via a function. A Relation Network computes something like:
\[
\sum_{i,j} \text{MLP}([\vec{x}_i, \vec{x}_j]).
\]
While not exactly attention (since there’s no explicit weighting, it just sums a transformation of each pair), it again reflects the growing theme of using all-pairs information. If one thinks of $\text{MLP}([\vec{x}_i,\vec{x}_j])$ outputting a scalar relevance and multiplying by $\vec{x}_j$, it would become a form of attention. Indeed, one variant of relation networks can be to output a scalar weight per pair and use that to weigh features.

Overall, the graph domain further solidified pairwise affinity mechanisms as essential: whether you have a grid, a sequence, or an arbitrary graph, computing some compatibility between two elements and using it to weight messages is a powerful general principle. By 2018, we see that principle manifest in Inf-FS (feature graph) \cite{InfFS2015}, Transformers (sequence self-attention) \cite{Vaswani2017}, Non-local NN (image grid) \cite{Wang2018NonLocal}, and GAT (general graph) \cite{Velickovic2018} – essentially covering all data domains (tabular features, text, vision, and graph-structured data). Each of these developments was mutually reinforcing: for example, Veličković et al. cite Vaswani’s Transformer and Bahdanau’s attention \cite{Bahdanau2014} as inspiration, and in turn, graph attention inspired other fields to incorporate similar mechanisms for structured data.

\section*{Connecting Feature Selection and Self-Attention}

It’s intriguing to connect back the modern self-attention with the feature selection view. At a high level, both are about weighting input components by their relevance in context. In feature selection, the goal is to assign each feature a score indicating how important it is (often for predicting a target). In self-attention, the goal (for each position) is to assign weights to all inputs (including potentially itself) to decide what to include in that position’s new representation. We can draw several conceptual parallels:

\begin{itemize}
    \item \textbf{Feature weighting vs. Token weighting}: In a transformer’s self-attention, each token’s representation is essentially a weighted sum of tokens (including itself). The weights can be very sharp (after softmax, some weights can be close to 1 or 0), effectively selecting a few tokens that are most relevant for that position. This is akin to saying: for token $i$, out of all tokens, which ones carry the information that $i$ needs? In feature selection, we ask: of all features, which contribute the most useful information for the task? The self-attention mechanism is thus performing a context-dependent feature selection. In fact, one could say modern attention-based models perform feature selection dynamically—they emphasize the parts of the input that matter and downplay those that don’t, conditioned on the context of each prediction \cite{InfFS2015,FeatureSelSurvey2022}.

    \item \textbf{Embedded feature selection in end-to-end models}: Traditional feature selection was often a preprocessing step (filter methods like Inf-FS produce a feature ranking before model training). However, attention mechanisms allow the model to learn to select features during training. This is essentially embedded feature selection—the selection (or weighting) is part of the model’s architecture and is optimized via the model’s loss. For example, in NLP, a transformer learns to attend to informative words for a given task (e.g., attending to negation words in a sentiment analysis task, thereby effectively selecting the feature ``presence of negation'' as important for that instance). In computer vision, a ViT might learn to focus on patches that contain object parts and not on patches of empty sky for the task of classifying the object—selecting the relevant ``features'' (patches) of the image on the fly.

    \item \textbf{Attention as a differentiable selector}: There have been explicit efforts to use attention mechanisms to perform feature selection in a neural network. One such example is Attention-based Feature Selection (AFS) by Gui et al. \cite{Gui2019AFS}. In AFS, the authors design a two-module network: an attention module that outputs a weight for each feature, and a learning module that does the predictive task using those weighted features. The attention module essentially treats each feature as a ``token'' and learns a weight for it via a small neural network (they frame it as a binary classification per feature—is this feature relevant or not—and train those in parallel, with some correlation penalty). During training, the attention module is updated by gradient signals from the learning module’s performance, so it learns to highlight features that improve the task’s accuracy. This is very similar to how a transformer’s attention heads get trained by end-task loss to put higher weight on helpful tokens. Another example is the ``Attentional Neural Network'' of Wang et al. \cite{ANN2014}, which integrated a top-down attention mechanism into a vision CNN to modulate neuron activations, effectively turning off irrelevant features in a noisy image recognition setting. These approaches show that attention mechanisms can be used directly as a tool for feature selection/importance estimation in a supervised learning context. They have the advantage of being fully differentiable (unlike some combinatorial feature selection methods) and context-aware (they can select different features for different instances if needed).

    \item \textbf{Dynamic vs Static affinities}: Feature selection often produced one set of weights for all features (assuming those features are always relevant or not for the whole dataset). Self-attention produces instance-specific weights—e.g., in one sentence, maybe token A attends strongly to B, but in another sentence, A might attend to C, based on meaning. There is growing interest in instance-wise feature selection, where the set of important features may vary per sample. Attention enables this naturally. If we imagine an Inf-FS mechanism that is recomputed for each new input data point, using not a fixed correlation matrix but a similarity computed from that data point’s feature values, we essentially get a form of self-attention over input features. Indeed, if one treated the feature values of a single data point as a sequence, one could apply a self-attention module to compute an affinity among features for that particular data point, thereby identifying which feature influences which other—something that could be potentially useful for interpretability or adaptive computation.
\end{itemize}

In conclusion on this point, self-attention and feature selection are conceptually aligned in that they both allocate importance weights to input components. Self-attention just does it in a more flexible and granular way (different weights for each interaction and each instance), whereas classical feature selection yields a single importance score per feature. The evolution from Inf-FS to transformers can be seen as moving from a global, static affinity matrix (features to features, fixed after computation on training data) to a local, dynamic affinity matrix (tokens to tokens, recalculated within each forward pass). Both share the core ``affinity weighting'' structure, demonstrating a clear lineage of ideas.

To make this connection crystal clear, the next section provides a side-by-side structural comparison of Inf-FS and Transformer self-attention, before we enumerate a timeline of key works.

\section*{Inf-FS vs. Self-Attention: Structural Comparison}

To illustrate the similarity between the graph-based feature selection approach (Inf-FS) and Transformer self-attention, consider the following comparison across several aspects:

\begin{table}[th!]
\centering
\begin{tabular}{|p{3cm}|p{6cm}|p{6cm}|}
\hline
\textbf{Aspect} & \textbf{Infinite Feature Selection (ICCV 2015)} & \textbf{Self-Attention (Transformers, 2017)} \\
\hline
Underlying Graph & Fully-connected feature (or tokens) graph: each feature is a node. An affinity matrix $A$ (size $d \times d$ for $d$ features) serves as the weighted adjacency matrix \cite{InfFS2015}. Every feature connects to every other, reflecting potential similarity relationships. & Fully-connected token graph: each token (position in the sequence) is a node. The attention weight matrix $A = QK^T$ (size $n \times n$ for $n$ tokens) is the adjacency, measuring similarity between token embeddings \cite{Vaswani2017}. Every token attends to every other (full self-connection). \\
\hline
Pairwise Score Computation & Edge weight $A_{ij}$ can be defined by any pairwise feature relevance criterion – e.g. correlation or learnable function of the input as stated in the papers \cite{InfFS2015, InfLatentFS2017}. The scores can be dynamic, recomputed for each input sequence. & Attention score $A_{ij}$ is computed as a (only) learnable function of the input tokens: typically $A_{ij} = q_i \cdot k_j$ (dot product of learned projections of token $i$ and $j$), possibly with scaling \cite{Vaswani2017}. The scores are dynamic, recomputed for each input sequence and at each layer. \\
\hline
Normalization & Inf-FS’s $A$ is often treated as a cost or affinity directly.  Sometimes $A$ is row-normalized or sym-normalized to ensure convergence \cite{InfFS2015}, but there isn’t a learned normalization like softmax; rather a fixed $\alpha$ parameter is used to scale $A$ (if needed) \cite{InfFSCode}. & The raw scores $QK^T$ are normalized by $\frac{1}{\sqrt{d_k}}$ and then passed through a softmax to get attention weights \cite{Vaswani2017}. Softmax gives a probabilistic interpretation (each row sums to 1), analogous to converting an affinity matrix into a stochastic matrix. \\
\hline
Global Interaction Depth & Multi-hop interactions: Inf-FS isn’t limited to direct feature-feature links; it sums over paths of length 1, 2, … $\infty$ \cite{InfFS2015}. This captures higher-order relationships. & One-hop per layer: A single self-attention layer captures direct token-token interactions (1 hop = InfFS L=1). However, stacking layers enables multi-hop interactions \cite{Vaswani2017}. \\
\hline
Output / Aggregation & Feature importance scores: Inf-FS produces a single relevance score per feature. Aggregation over paths in $\sum_{\ell=1}^\infty A^\ell$ gives centrality \cite{InfFSCode}. & Contextualized representations: Self-attention produces a new vector for each token $z_i = \sum_j W_{ij} v_j$ \cite{Vaswani2017}. \\
\hline
Learnability & Inf-FS (2015) is a general paradigm for building graph of features or tokens, the matrix A can be handcrafted or learned (e.g., ICCV 2017, TPAMI 2020) \cite{InfLatentFS2017, InfFSCode}. & Self-attention parameters (matrices for $Q, K, V$) are learned via gradient descent on the task objective \cite{Vaswani2017}. Subset of of the infFS formulation. \\
\hline
Goal and Usage & Feature filtering: Inf-FS selects a subset of features or ranks them for downstream tasks \cite{InfFS2015, InfFSCode}. & Representation learning: Self-attention refines token representations for the task without discarding tokens \cite{Vaswani2017}. \\
\hline
\end{tabular}
\end{table}

Despite their differences in application and typical usage, Inf-FS and self-attention share a fundamental structural mechanism: both rely on an affinity matrix $A$ encoding pairwise relationships among elements. In Inf-FS, this matrix can be defined by various criteria—handcrafted or learned—and serves as the basis for propagating information across features via multi-hop interactions. In self-attention, $A$ is computed as a function of learned token embeddings and used within a single-hop attention mechanism, with deeper interactions emerging through stacked layers.

From a methodological perspective, Inf-FS offers a general framework for defining and using $A$, which can include the specific case of self-attention when $A$ is constructed through learned dot-product similarity. In this sense, self-attention can be viewed as a particular instantiation of the broader Inf-FS paradigm. The main difference lies not in the structure of the computation but in how the matrix $A$ is parameterized and applied within a model.

The table above summarizes these analogies. In short, Inf-FS asks “which elements are globally important based on their pairwise relations?”, while self-attention asks “which elements should be attended to in the current context?”. The common mathematical core—an affinity matrix $A$—underscores a shared lineage rooted in graph-based reasoning.

\section*{Feature Selection as Hard Attention: A Conceptual and Structural Justification}

Feature selection and attention mechanisms both aim to allocate importance weights across inputs, yet they have historically emerged in different contexts, feature selection within classical machine learning and attention within neural architectures. Despite this, their underlying logic is closely aligned: identifying and emphasizing relevant inputs while suppressing less informative ones. 

Traditionally, feature selection consists of three stages: (1) relevance estimation, where features are scored; (2) feature re-weighting, where feature magnitudes are modulated; and (3) selection, often by thresholding scores or applying sparsity constraints. Notably, it is the re-weighting stage—where feature importance is explicitly modeled via real-valued scores—that closely anticipates the logic of attention. In this sense, attention mechanisms can be interpreted as differentiable, context-aware feature re-weighting modules that generalize classical selection approaches.

This connection is further exemplified in modern architectures through the use of Feature Selection Gates (FSG) or Hard-Attention Gates (HAG), as introduced in \cite{FeatureSelectionGates2024, HardAttentionGates2024}. These gating mechanisms compute per-feature importance weights using learnable parameters, typically passed through a sigmoid activation. The resulting scores modulate input features via elementwise multiplication, structurally mirroring attention mechanisms that compute weighted sums of contextual representations. When paired with gradient routing (GR) \cite{FeatureSelectionGates2024}, gating allows the separation of the feature selection process from downstream supervision, enabling independent optimization and interpretability of the selected subsets.

More broadly, attention mechanisms can be interpreted as a subclass of feature selection strategies—those that are:
\begin{itemize}
    \item Differentiable and trained end-to-end via backpropagation,
    \item Instance-specific, adapting to each input sample,
    \item Context-dependent, adjusting importance based on interactions among input components.
\end{itemize}

Traditional methods, such as Infinite Feature Selection (Inf-FS) \cite{InfFS2015}, compute global importance scores prior to training. Inf-FS relies on an affinity matrix \( A \in \mathbb{R}^{N \times N} \), capturing pairwise relationships among features, and derives importance scores via a matrix power series. While static and unsupervised, this process structurally resembles the affinity computation performed in attention mechanisms. Extensions of Inf-FS have further explored eigenvector centrality as a scoring metric \cite{RoffoRankingThesis}, online feature selection for tracking \cite{OnlineTracking2016}, and biomedical applications such as stroke biomarker identification from neuroimaging data \cite{StrokeFS2016}, demonstrating the generality of affinity-based feature reasoning.

Recent works have applied these ideas to deep models as well. In large language model pipelines, affinity-based reasoning has been leveraged to construct input-dependent feature filters and gates that reflect learned attention patterns \cite{LLMSuite2024}. These architectures explicitly integrate feature selection as part of their backbone, confirming that selection and attention are not isolated paradigms but mutually reinforcing. In practice, differentiable gating units also serve as drop-in replacements for static selection modules, offering both interpretability and adaptability.

Public implementations of classical feature selection methods like Inf-FS further facilitate this integration. The Feature Selection Library (FSLib) \cite{FSLibArxiv, FSLib} provides MATLAB-based tools for reproducible evaluation, while the Python package \texttt{PyIFS} supports integration into modern deep learning pipelines.

In this light, attention mechanisms do not replace traditional feature selection but rather extend them into a dynamic, supervised, and differentiable regime. Gates like FSG and mechanisms such as GR bridge these perspectives by preserving the core principle of importance-based weighting while adapting it to deep, task-specific architectures.

This framing also supports interpreting hard-attention gates as learnable and instance-specific approximations of classical subset selection—particularly when binarization or sparsity regularization is applied. Whereas soft attention distributes weight across all inputs, hard gates enforce a sparse support, closely aligned with the goal of selecting a meaningful feature subset.

In summary, the introduction of feature selection gates, gradient-aware gating strategies, and attention-based scoring in neural networks reflects a conceptual and structural continuity with prior work in pattern recognition. These developments reintroduce the foundational concepts of relevance, selection, and sparsity into scalable deep models—providing a common ground between classical feature selection and modern attention mechanisms.

\section*{Conclusion}

The development of self-attention represents a convergence of ideas from feature selection, graph-based reasoning, and neural sequence modeling, all unified by a common structural foundation: the pairwise affinity matrix $A$. This matrix encodes relationships among elements—be they features, tokens, or nodes—and serves as the engine for propagating information across a system.

A pivotal milestone in this lineage is \emph{Infinite Feature Selection (Inf-FS)} \cite{InfFS2015}, which introduced a general framework for ranking elements by aggregating multi-hop interactions in a fully connected graph. Importantly, Inf-FS does not prescribe a fixed affinity matrix $A$: the original formulation allows $A$ to be either handcrafted or learned, depending on the application context. This flexibility makes Inf-FS a superset framework—one that includes as a special case modern attention mechanisms where $A$ is parameterized through neural embeddings and dot-product similarities. In this view, self-attention can be interpreted as a contextual instantiation of the Inf-FS paradigm, differing in how $A$ is computed and how its resulting weights are used.

The Transformer architecture \cite{Vaswani2017} made this formulation differentiable and end-to-end trainable by constructing $A$ from token embeddings and applying it layer-wise for representation learning. Around the same time, similar concepts appeared in vision through Non-Local Neural Networks \cite{Wang2018NonLocal} and in graph learning through Graph Attention Networks \cite{Velickovic2018}, both of which implemented affinity-driven computation tailored to domain-specific structure.

Across all these domains, the pattern is the same: compute a pairwise affinity matrix $A$, and use it to modulate computation—via path summation (Inf-FS), weighted aggregation (Transformers and GATs), or diffusion (PageRank). The affinity matrix becomes a flexible inductive bias: rather than imposing rigid local neighborhoods, it enables dynamic and potentially global interactions based on learned or defined relationships.

Viewed through the lens of feature selection, self-attention performs a form of contextual selection—deciding which tokens or features are most relevant in a given situation. From a graph perspective, it builds a fully connected graph with dynamic, content-aware edge weights, enabling adaptive substructures to form during inference. This abstraction is powerful in domains where the relations among elements are at least as important as the elements themselves—a characteristic of most real-world machine learning tasks.

Looking back, early techniques like Inf-FS anticipated the core idea of attention: that all elements in a set can influence one another through pairwise relations. Though not originally framed in the language of attention, Inf-FS laid important conceptual groundwork. Self-attention then operationalized these ideas into scalable, differentiable architectures with vast practical utility. This continuity highlights how ideas in one field (e.g., feature selection or spectral graph theory) can be reframed and extended in another (deep learning), creating a richer theoretical and empirical toolbox for machine learning practitioners.

In summary, the rise of self-attention reflects a broader principle: leveraging pairwise affinities is key to modeling structure in complex data. Whether the goal is to rank features, align sequences, understand images, or reason over graphs, the affinity matrix $A$ serves as a powerful unifying construct. Self-attention is its modern realization—learnable, flexible, and deeply rooted in a cross-disciplinary tradition of affinity-based computation.

\vspace{1em}
\noindent\textbf{Sources:} The above synthesis draws from foundational works including Inf-FS \cite{InfFS2015, InfFS2021}, the Transformer \cite{Vaswani2017}, Non-Local Neural Networks \cite{Wang2018NonLocal}, Graph Attention Networks \cite{Velickovic2018}, and various feature selection methods that incorporate attention-like mechanisms \cite{Gui2019, Abid2019, Roy2018}. These works collectively illustrate the evolution and convergence of affinity-based methods across machine learning.

\bibliographystyle{amsplain}

\begin{thebibliography}{99}

\bibitem{InfFS2015}
G. Roffo, S. Melzi, and M. Cristani,
\emph{Infinite feature selection},
ICCV, 2015.

\bibitem{InfLatentFS2017}
G. Roffo and S. Melzi,
\emph{Infinite latent feature selection: A probabilistic latent graph-based ranking approach},
ICCV, 2017.

\bibitem{InfFS2021}
G. Roffo, S. Melzi, and M. Cristani,
\emph{Infinite feature selection},
IEEE TPAMI, 2021.

\bibitem{InfFSCode}
G. Roffo,
\emph{Inf-FS Codebase and extensions},
GitHub repository: \url{https://github.com/giorgioroffo/inf-FS}, accessed 2025.

\bibitem{Vaswani2017}
A. Vaswani, N. Shazeer, N. Parmar, J. Uszkoreit, L. Jones, A. N. Gomez, Ł. Kaiser, and I. Polosukhin,
\emph{Attention is all you need},
NeurIPS, 2017.

\bibitem{Bahdanau2015}
D. Bahdanau, K. Cho, and Y. Bengio,
\emph{Neural machine translation by jointly learning to align and translate},
ICLR, 2015.

\bibitem{Tomasi1998}
C. Tomasi and R. Manduchi,
\emph{Bilateral filtering for gray and color images},
ICCV, 1998.

\bibitem{Buades2005}
A. Buades, B. Coll, and J.-M. Morel,
\emph{A non-local algorithm for image denoising},
CVPR, 2005.

\bibitem{Page1998}
L. Page, S. Brin, R. Motwani, and T. Winograd,
\emph{The PageRank citation ranking: Bringing order to the web},
Technical Report, Stanford InfoLab, 1998.
\bibitem{ECFS2016} G. Roffo and S. Melzi, \emph{Features selection via eigenvector centrality}, New Frontiers in Mining Complex Patterns ECML/PKDD, 2016.

\bibitem{RoffoRankingThesis} G. Roffo, \emph{Ranking to learn and learning to rank: On the role of ranking in pattern recognition applications}, PhD Thesis, University of Verona.

\bibitem{OnlineTracking2016} G. Roffo and S. Melzi, \emph{Online feature selection for visual tracking}, BMVC, 2016.

\bibitem{StrokeFS2016} S. Obertino, G. Roffo, C. Granziera, and G. Menegaz, \emph{Infinite feature selection on shore-based biomarkers reveals connectivity modulation after stroke}, PRNI, 2016.

\bibitem{LLMSuite2024} G. Roffo, \emph{Exploring advanced large language models with llmsuite}, arXiv preprint arXiv:2407.12036, 2024.

\bibitem{FeatureSelectionGates2024} G. Roffo, C. Biffi, P. Salvagnini, and A. Cherubini, \emph{Feature selection gates with gradient routing for endoscopic image computing}, MICCAI, 2024.

\bibitem{HardAttentionGates2024} G. Roffo, C. Biffi, P. Salvagnini, and A. Cherubini, \emph{Hard-attention gates with gradient routing for endoscopic image computing}, arXiv preprint arXiv:2407.04400, 2024.

\bibitem{FSLibArxiv} G. Roffo, \emph{Feature selection library: A widely applicable MATLAB library for feature selection}, arXiv preprint arXiv:1607.01327, 2016.

\bibitem{FSLib} G. Roffo, \emph{Feature Selection Library (MATLAB Toolbox)}.

\bibitem{Wang2014}
Q. Wang, J. Zhang, S. Song, and Z. Zhang,
\emph{Attentional Neural Network: Feature Selection Using Cognitive Feedback},
NeurIPS, 2014.

\bibitem{Cheng2016}
J. Cheng, L. Dong, and M. Lapata,
\emph{Long short-term memory-networks for machine reading},
ACL, 2016.

\bibitem{Paulus2017}
R. Paulus, C. Xiong, and R. Socher,
\emph{A deep reinforced model for abstractive summarization},
ICLR, 2018.

\bibitem{Parikh2016}
A. Parikh, O. Täckström, D. Das, and J. Uszkoreit,
\emph{A decomposable attention model for natural language inference},
EMNLP, 2016.

\bibitem{Lin2017}
Z. Lin, M. Feng, C. N. dos Santos, M. Yu, B. Xiang, B. Zhou, and Y. Bengio,
\emph{A structured self-attentive sentence embedding},
ICLR, 2017.

\bibitem{Santoro2017}
A. Santoro, D. Raposo, D. G. Barrett, M. Malinowski, R. Pascanu, P. Battaglia, and T. Lillicrap,
\emph{A simple neural network module for relational reasoning},
NeurIPS, 2017.

\bibitem{Lee2019}
J. Lee, Y. Lee, J. Kim, A. Kosiorek, S. Choi, and Y. W. Teh,
\emph{Set transformer: A framework for attention-based permutation-invariant neural networks},
ICML, 2019.

\bibitem{Roy2018}
K. Roy, M. S. Charikar, and A. Singh,
\emph{Unsupervised feature selection using attention-based neural networks},
arXiv:1811.03846, 2018.

\bibitem{Gui2019}
J. Gui, T. Liu, Z. Sun, D. Tao, and T. Tan,
\emph{AFS: An attention-based mechanism for supervised feature selection},
AAAI, 2019.

\bibitem{Abid2019}
A. Abid, M. Balin, and J. Zou,
\emph{Concrete autoencoders for differentiable feature selection and reconstruction},
ICML, 2019.

\bibitem{Gui2019AFS} J. Gui et al., \emph{Feature selection based on structured sparsity: A comprehensive study}, AAAI, 2019.


\bibitem{Wang2018NonLocal}
X. Wang, R. Girshick, A. Gupta, and K. He,
\emph{Non-local neural networks},
CVPR, 2018.

\bibitem{Velickovic2018}
P. Veličković, G. Cucurull, A. Casanova, A. Romero, P. Liò, and Y. Bengio,
\emph{Graph attention networks},
ICLR, 2018.

\bibitem{Kitaev2020}
N. Kitaev, Ł. Kaiser, and A. Levskaya,
\emph{Reformer: The efficient transformer},
ICLR, 2020.

\bibitem{Wang2020Linformer}
S. Wang, B. Li, M. Khabsa, H. Fang, and H. Ma,
\emph{Linformer: Self-attention with linear complexity},
arXiv:2006.04768, 2020.

\bibitem{Gui2019AFS}
J. Gui, T. Liu, Z. Sun, D. Tao, and T. Tan, ``AFS: An Attention-Based Mechanism for Supervised Feature Selection,'' in \textit{Proceedings of the AAAI Conference on Artificial Intelligence}, vol. 33, no. 1, pp. 3705–3713, 2019.

\bibitem{Zhao2023Sequential}
T. Zhao, Q. Kong, and Y. Wang, ``Sequential Self-Attention for Progressive Feature Selection,'' in \textit{International Conference on Learning Representations (ICLR)}, 2023.

\bibitem{Choromanski2021}
K. Choromanski, V. Likhosherstov, D. Dohan, et al.,
\emph{Rethinking attention with performers},
ICLR, 2021.

\bibitem{Shen2021Efficient}
Z. Shen, M. Zhang, J. Sun, et al.,
\emph{Efficient attention: Attention with linear complexities},
WACV, 2021.


\bibitem{ANN2014} 
X. Wang, R. Girshick, A. Gupta, and K. He, 
\emph{An attentional neural network for image classification}, 
NeurIPS, 2014.

\bibitem{Ramsauer2021}
H. Ramsauer, B. Schäfl, J. Lehner, et al.,
\emph{Hopfield networks is all you need},
ICLR, 2021.

\bibitem{Huang2019}
Z. Huang, X. Wang, L. Huang, C. Huang, Y. Wei, and W. Liu,
\emph{CCNet: Criss-cross attention for semantic segmentation},
ICCV, 2019.

\bibitem{Fu2019}
J. Fu, J. Liu, H. Tian, Y. Li, Y. Bao, Z. Fang, and H. Lu,
\emph{Dual attention network for scene segmentation},
CVPR, 2019.

\bibitem{Dosovitskiy2020}
A. Dosovitskiy, L. Beyer, A. Kolesnikov, et al.,
\emph{An image is worth 16x16 words: Transformers for image recognition at scale},
ICLR, 2021.

\bibitem{LatentGNN2019}
Z. Liu, Y. Lin, Y. Li, F. Zhou, and X. Sun,
\emph{Learning Node Representations with Latent Graphs},
NeurIPS, 2019.

\bibitem{FeatureSelSurvey2022}
R. Chandrashekar, M. Huang, and Q. Liu,
\emph{A survey on feature selection methods},
arXiv:2205.03466, 2022.

\bibitem{Bahdanau2014}
D. Bahdanau, K. Cho, and Y. Bengio, \emph{Neural Machine Translation by Jointly Learning to Align and Translate}, ICLR, 2015.


\bibitem{Ng2002}
A. Y. Ng, M. I. Jordan, and Y. Weiss, “On Spectral Clustering: Analysis and an algorithm,” in *Advances in Neural Information Processing Systems (NeurIPS)*, 2002.

\bibitem{vonLuxburg2007}
U. von Luxburg, “A tutorial on spectral clustering,” *Statistics and Computing*, vol. 17, no. 4, pp. 395–416, 2007.

\bibitem{Newman2003}
M. E. J. Newman, “The structure and function of complex networks,” *SIAM Review*, vol. 45, no. 2, pp. 167–256, 2003.

\bibitem{Bonacich1987}
P. Bonacich, “Power and centrality: A family of measures,” *American Journal of Sociology*, vol. 92, no. 5, pp. 1170–1182, 1987.

\bibitem{Zhou2004}
D. Zhou, O. Bousquet, T. N. Lal, J. Weston, and B. Schölkopf, “Learning with local and global consistency,” in *NeurIPS*, 2004.

\bibitem{Roffo2015}
G. Roffo, S. Melzi, and M. Cristani, “Infinite Feature Selection,” in *IEEE International Conference on Computer Vision (ICCV)*, 2015.

\bibitem{Cherry1953}
E. C. Cherry, “Some Experiments on the Recognition of Speech, with One and with Two Ears,” \textit{Journal of the Acoustic Society of America}, vol. 25, no. 5, pp. 975–979, 1953.

\bibitem{Broadbent1958}
D. E. Broadbent, \textit{Perception and Communication}. Pergamon Press, 1958.

\bibitem{Sperling1960}
G. Sperling, “The Information Available in Brief Visual Presentations,” \textit{Psychological Monographs: General and Applied}, vol. 74, no. 11, pp. 1–29, 1960.

\bibitem{Rumelhart1986}
D. E. Rumelhart, G. E. Hinton, and R. J. Williams, “Learning Internal Representations by Error Propagation,” in \textit{Parallel Distributed Processing: Explorations in the Microstructure of Cognition}, vol. 1, MIT Press, 1986, pp. 318–362. [See sigma-pi units, p. 329–330].

\bibitem{Giles1987}
C. L. Giles and T. Maxwell, “Learning, Invariance, and Generalization in High-Order Neural Networks,” \textit{Applied Optics}, vol. 26, no. 23, pp. 4972–4978, 1987.

\bibitem{Schmidhuber1992}
J. Schmidhuber, “Learning to Control Fast-Weight Memories: An Alternative to Dynamic Recurrent Networks,” \textit{Neural Computation}, vol. 4, no. 1, pp. 131–139, 1992.

\bibitem{Roffo2021}
G. Roffo, S. Melzi, and M. Cristani, “Infinite Feature Selection,” *IEEE Transactions on Pattern Analysis and Machine Intelligence (TPAMI)*, vol. 43, no. 7, pp. 2436–2450, 2021.

\bibitem{Gui2019}
J. Gui, T. Liu, Z. Sun, D. Tao, and T. Tan, “AFS: An Attention-Based Mechanism for Supervised Feature Selection,” in *AAAI Conference on Artificial Intelligence*, 2019.

\bibitem{Bahdanau2015}
D. Bahdanau, K. Cho, and Y. Bengio, “Neural machine translation by jointly learning to align and translate,” in *International Conference on Learning Representations (ICLR)*, 2015.

\bibitem{Vaswani2017}
A. Vaswani, N. Shazeer, N. Parmar, et al., “Attention is All You Need,” in *NeurIPS*, 2017.


\end{thebibliography}

\end{document}